\documentclass[10pt,twocolumn,letterpaper]{article}
\usepackage{scrid-preprint}

\newcommand{\AuthorNames}{%
  Hao Tan\textsuperscript{1,2}\quad
  Zekai Zuo\textsuperscript{1}\quad
  Boyang Chen\textsuperscript{1}\quad
  Jinghui Qin\textsuperscript{1,*}%
}
\newcommand{\AuthorAffiliations}{%
  \textsuperscript{1}Guangdong University of Technology\\[1pt]
  \textsuperscript{2}The Hong Kong University of Science and Technology (Guangzhou)%
}
\newcommand{\AuthorEmails}{%
  tanhao4869@gmail.com\quad zuozekai@mails.gdut.edu.cn\\[1pt]
  chenboyang2@mails.gdut.edu.cn\quad scape1989@gmail.com%
}

\makeatletter
\renewcommand{\maketitle}{%
  \twocolumn[{%
    \begin{center}
      {\fontsize{18}{21}\selectfont\bfseries\@title\par}
      \vspace{12pt}
      {\fontsize{12}{14}\selectfont\bfseries\AuthorNames\par}
      \vspace{4pt}
      {\fontsize{11}{13}\selectfont\AuthorAffiliations\par}
      \vspace{3pt}
      {\fontsize{9}{11}\selectfont\ttfamily\AuthorEmails\par}
    \end{center}
    \vspace{7pt}
  }]
  \begingroup
    \renewcommand{\thefootnote}{\fnsymbol{footnote}}
    \footnotetext[1]{Corresponding author.}
  \endgroup
  \setcounter{footnote}{0}
}
\makeatother

\title{SCRiD: Spatial and Contextual References for Generalizable AI-Generated Video Detection}
\author{Hao Tan \and Zekai Zuo \and Boyang Chen \and Jinghui Qin}
\date{}

\hypersetup{
  pdftitle={SCRiD: Spatial and Contextual References for Generalizable AI-Generated Video Detection},
  pdfauthor={Hao Tan, Zekai Zuo, Boyang Chen, Jinghui Qin}
}

\begin{document}
\maketitle

\begin{abstract}
Generalizable AI-generated video detection remains challenging because natural motion and content variations can resemble synthesis artifacts, making isolated video discrepancies unreliable indicators of authenticity. To address this challenge, we propose Spatial and Contextual Reference-informed Detection (SCRiD), a framework that interprets video variations through complementary spatial and contextual references. Specifically, we introduce Matched-Coordinate Residual Response (MCR) to characterize temporal residual associations under motion and coordinate references with matched subpixel sampling conditions. We further develop Real-Conditioned Score Adaptation (RCSA), which learns a content-conditioned denoising reference from authentic videos and probes its responses across multiple noise scales and quality views to guide feature adaptation. To integrate these complementary sources of evidence, we design Reversible Evidence Exchange (REE) to enable information exchange through reversible symplectic updates. 
We train SCRiD using synthetic videos from only 4 generators and evaluate it across 88 generator or task subsets spanning 6 benchmarks. Our method achieves average ACC and AP of 72.72\% and 74.23\%, respectively, outperforming the strongest baseline by 3.24 and 0.81 percentage points.
Ablation and robustness experiments further validate the effectiveness of the proposed components and the stability of SCRiD under common post-processing operations.
\end{abstract}
\section{Introduction}
\label{sec:intro}


\begin{figure}[t]
\centering
\includegraphics[width=\columnwidth]{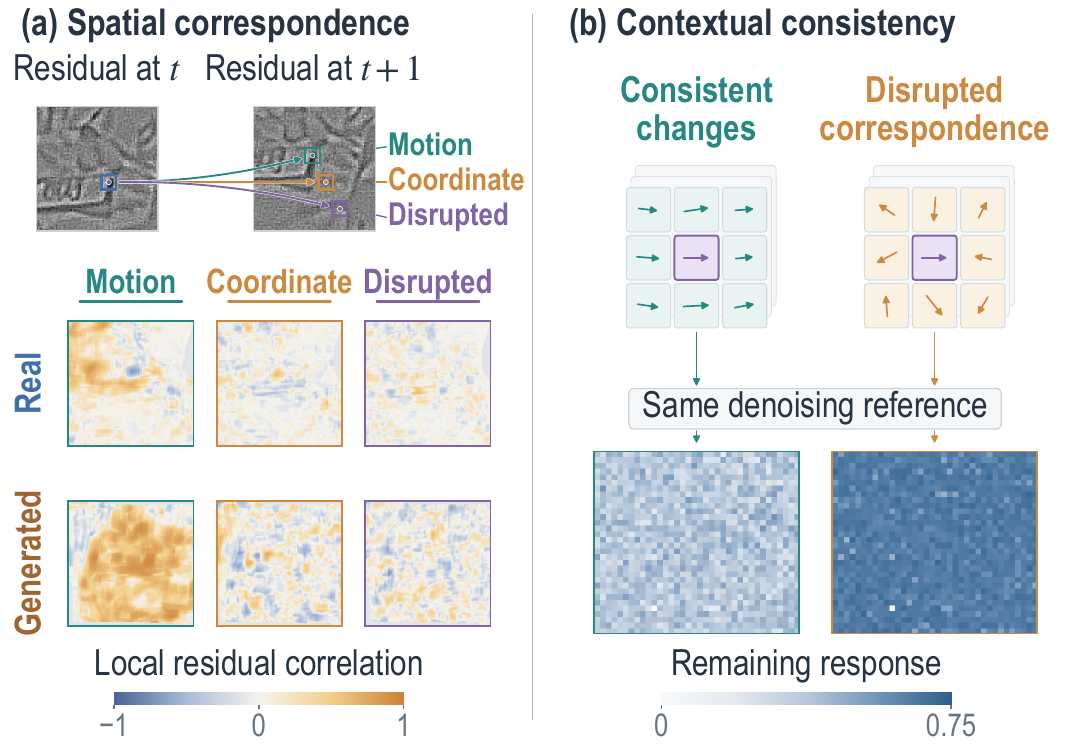}
\caption{Motivating observations on spatial correspondence and contextual consistency. (a) Motion correspondence reveals clearer residual correlation differences between real and generated samples than coordinate-based comparisons. (b) With local feature variations fixed, consistent contextual changes produce smaller denoising responses than disrupted ones.}
\label{fig:intro}
\end{figure}

Modern video generation models produce increasingly realistic synthetic videos~\cite{blattmann2023stable, yang2025cogvideox}, offering substantial creative potential. However, they also make it hard for humans to verify the authenticity of visual media, posing serious security risks, such as misinformation spread, public opinion manipulation, etc. Consequently, developing reliable, generalizable AI-generated video detectors has become highly urgent yet challenging, since generalizable AI-generated video detection requires a detector trained on samples from a limited set of generators to recognize videos produced by unseen ones~\cite{yang2026svd}. Existing approaches have explored various sources of detection evidence. Frame-level methods~\cite{qian2020thinking, tan2024rethinking} identify synthesis-related spatial and frequency artifacts or leverage pretrained visual encoders to extract transferable representations~\cite{ojha2023towards, bi2026radi}. Video-level methods~\cite{ma2024decof, chen2026demamba, zheng2025d3} further exploit temporal information by analyzing inter-frame differences, motion patterns, and higher-order dynamics. Although these advances have broadened the range of cues available for distinguishing real and generated videos, such methods also capture natural variations in authentic videos, making synthesis-related discrepancies harder to distinguish when facing unseen generators.


Natural video often contains rapid motion, camera movements, rich textures, and changing illumination, all of which can produce pronounced residuals and feature variations without any synthesis~\cite{liu2025research, song2025vf}. Meanwhile, generated videos may exhibit plausible appearances and smooth temporal transitions, leaving only subtle irregularities that are difficult to recognize~\cite{jiang2026ivy}. As a result, a large discrepancy does not necessarily indicate synthesis, while a small discrepancy does not guarantee authenticity. Moreover, synthesis artifacts may vary considerably across generators~\cite{ma2026your}, limiting the reliability of patterns learned from known generation methods. These difficulties suggest that the significance of an observed discrepancy cannot be determined by its magnitude alone, but also depends on how it relates to natural changes in video content. Distinguishing such natural variations from synthesis-related irregularities remains a central challenge for generalized video detection.

As objects move across frames, their residual patterns shift accordingly, making comparisons at fixed image locations sensitive to natural motion. 
As shown in Figure~\ref{fig:intro} (a), we observe that the same residual signals from paired generated and real videos exhibit different discriminative responses when compared at motion-corresponding locations and at nearby positions in the initial image coordinates. Moreover, these discriminative cues become weaker when the motion correspondence is disrupted. These observations indicate that the discriminative value of temporal residuals depends not only on their magnitude, but also on the spatial correspondence used to associate them across frames. 
Moreover, spatial correspondence only tells us where to compare residuals, leaving open whether the observed changes are consistent with the surrounding video content. A local feature may change naturally alongside related changes in other regions and frames, making it difficult to judge its significance in isolation. As illustrated in Figure~\ref{fig:intro} (b), we observe that even when the local feature variation is held fixed, the denoising response depends on the surrounding content: when changes in surrounding regions and frames are consistent with the local variation, the remaining denoising response is smaller, whereas disrupting this relationship leads to a larger response. This observation indicates that the significance of local feature variations also depends on how well they agree with changes in the surrounding content. Together, these findings suggest that identifying synthesis-related discrepancies requires considering both the spatial correspondence of temporal residuals and the contextual consistency of feature variations.

Motivated by these findings, we propose Spatial and Contextual Reference-informed Detection (SCRiD), a framework that identifies synthesis-related irregularities by examining how video changes behave under complementary reference conditions. Specifically, we introduce Matched-Coordinate Residual Response (MCR) to compare inter-frame residual associations along estimated motion and at nearby positions in the initial image coordinates. By matching their subpixel sampling conditions and measuring relative correlation and distance responses, we extract local temporal evidence that reflects inter-frame spatial correspondence. To further account for content-dependent variations, we develop Real-Conditioned Score Adaptation (RCSA) to learn a content-conditioned denoising reference from authentic videos. We then probe this reference through symmetric perturbations across multiple noise levels and quality views and use the resulting responses to adapt pretrained visual representations. Since these two components provide different types of evidence, we introduce Reversible Evidence Exchange (REE) to enable their interaction before fusion. Specifically, we maintain the two evidence streams as coupled states and exchange information through reversible, symplectic updates which can preserve the recoverability of their joint representation throughout the interaction stage. The exchanged representations are subsequently fused and aggregated for video-level classification. 
Extensive experiments on 6 benchmarks, including GenVidBench~\cite{ni2026genvidbench}, AIGVDBench~\cite{ma2026your}, ComGenVid~\cite{ben2026training}, GenBuster~\cite{wen2025busterx}, ViF-Bench~\cite{li2026skyra}, and RA-Bench~\cite{liang2026can}, demonstrate that SCRiD achieves state-of-the-art performance across diverse datasets and generation sources when trained exclusively on GenVidBench Pair1. 

Overall, our main contributions are summarized as follows:
\begin{itemize}
    \item We highlight the importance of spatial correspondence and contextual consistency in interpreting video discrepancies, providing a new perspective for generalized AI-generated video detection.
    \item We propose SCRiD, a reference-informed detection framework that combines matched-coordinate residual evidence with real-conditioned denoising responses and enables their interaction through reversible evidence exchange.
    \item Extensive experiments across 6 benchmarks demonstrate the strong generalization capability of SCRiD, outperforming existing detectors on diverse unseen generators and datasets.
\end{itemize}

\section{Related work}
\label{sec:rw}

\textbf{AI-Generated Video Detection.} AI-generated video detection has evolved from frame-level synthesis trace analysis~\cite{kundu2025towards,tan2024rethinking,ojha2023towards} toward explicit temporal modeling. DeCoF~\cite{ma2024decof} examines frame consistency, DeMamba~\cite{chen2026demamba} models spatial and temporal inconsistencies, and D3~\cite{zheng2025d3} exploits higher-order temporal variations without training. However, generalization to unseen generators remains challenging, as natural motion and content variations can resemble synthesis artifacts~\cite{vahdati2024beyond}. Recent methods address this challenge through forensic augmentation~\cite{corvi2026seeing} or spatial and temporal likelihoods derived from real videos~\cite{ben2026training}. Despite these advances, distinguishing synthesis artifacts from natural video variations remains difficult. Our SCRiD complements these approaches by interpreting video discrepancies under spatial and contextual references and integrating complementary evidence for generalizable detection.

\noindent\textbf{Temporal Consistency Modeling.} Analyzing temporal consistency helps detect video forgeries by revealing artifacts that may be overlooked when frames are examined in isolation.
STIL~\cite{gu2021spatiotemporal} captures spatiotemporal inconsistencies through adjacent-frame differences. FTCN~\cite{zheng2021exploring} models temporal coherence using convolutional features. AltFreezing~\cite{wang2023altfreezing} encourages the learning of both spatial and temporal artifacts. Kim et al.~\cite{kim2025beyond} explore pixel-wise temporal frequency patterns to identify subtle manipulations. Recent studies extend temporal analysis to fully AI-generated videos. ReStraV~\cite{interno2026ai} characterizes temporal trajectories in pretrained feature spaces, while MotionPhys~\cite{he2026motionphys} investigates the physical consistency of optical-flow trajectories. These studies highlight the importance of temporal dynamics in detecting synthetic content. 
Our SCRiD examines residual associations under motion-aligned and image-coordinate references to characterize how spatial correspondence affects the discriminative value of temporal residuals.

\noindent\textbf{Visual Prior Representation.} Pretrained visual encoders provide transferable representations that can reduce dependence on generator-specific artifacts. DINOv2~\cite{oquab2023dinov2} learns general-purpose visual features through self-supervised pretraining, while RealForensics~\cite{haliassos2022leveraging} leverages authentic videos to learn representations for robust forgery detection. Another line of research exploits reconstruction-based cues to distinguish real and generated content. DIRE~\cite{wang2023dire} measures reconstruction errors from pretrained diffusion models, whereas AEROBLADE~\cite{ricker2024aeroblade} examines the reconstruction behavior of latent-diffusion autoencoders. Beyond reconstruction, denoising score matching~\cite{vincent2011connection,song2019generative} offers a principled way to characterize data distributions through noise-dependent responses. These studies highlight the potential of pretrained representations and learned data priors for visual analysis. Building on these ideas, our SCRiD learns a content-conditioned denoising reference from authentic videos and probes its responses across multiple noise levels and quality views. The resulting responses guide representation adaptation to provide contextual evidence that complements temporal residual information.
\section{Method}
\label{sec:method}



\begin{figure*}[t]
\centering
\includegraphics[width=0.99\textwidth]{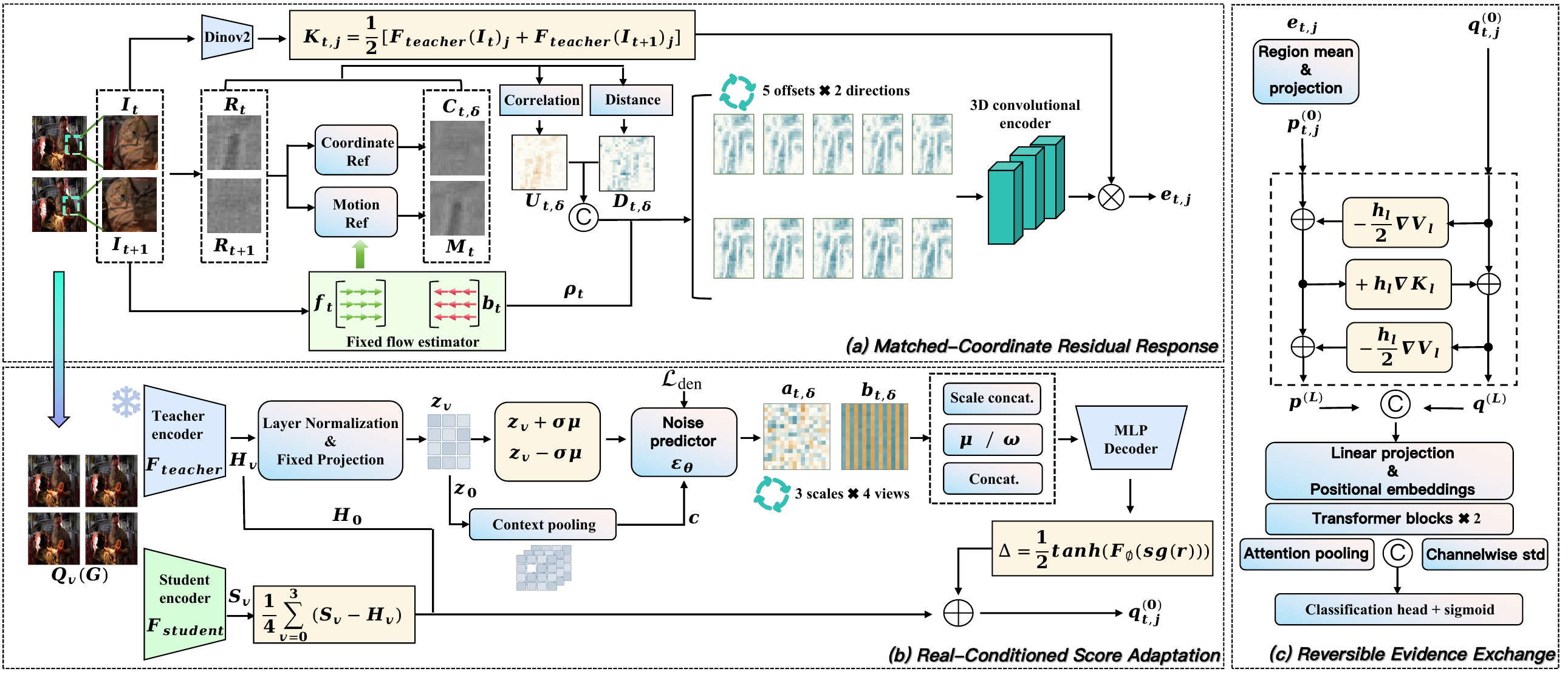}
\caption{Overview of the SCRiD framework. (a) MCR extracts temporal residual evidence using motion and coordinate references with matched subpixel sampling. (b) RCSA adapts visual features through real-conditioned denoising responses across noise scales and quality views. (c) REE integrates both evidence streams through reversible symplectic updates for video-level classification.}
\label{fig:model}
\end{figure*}


Given an input video $X=\{x_t\}^T_{t=1}$ with $T$ frames, SCRiD aims to distinguish authentic from AI-generated videos and generalize to unseen generators. SCRiD approaches this goal by interpreting video variations under two complementary references: 1) spatial correspondence, which determines where temporal residuals should be compared, and 2) content-dependent temporal consistency, which characterizes how local feature variations relate to their surrounding context. These references are modeled by Matched-Coordinate Residual Response (MCR) and Real-Conditioned Score Adaptation (RCSA), respectively. Their resulting representations are subsequently integrated through Reversible Evidence Exchange (REE) for video-level prediction. The overall architecture of SCRiD is illustrated in Figure.~\ref{fig:model}.

\subsection{Matched-Coordinate Residual Response}
Motivated by Figure.~\ref{fig:intro}(a), MCR compares residual associations under estimated motion and nearby image coordinates. We centrally crop each frame to obtain local views $\{I_t\}^T_{t=1}$ and extract residuals at two spatial scales:
\begin{equation}
    R_t=[I_t-\mathcal{A}_3(I_t);\mathcal{A}_3(I_t)-\mathcal{A}_7(I_t)].
\end{equation}
Here, $\mathcal A_k$ is a $k\times k$ average filter with reflection padding, and $[\cdot;\cdot]$ denotes channel-wise concatenation. A fixed estimator computes forward optical flow $f_t$ from $I_t$ to $I_{t+1}$. With bilinear sampling defined as $\mathcal W(S,f)(\xi)=S(\xi+f(\xi))$ for a spatial field $S$, displacement field $f$, and pixel coordinate $\xi$, we construct motion and nearby coordinate references:
\begin{equation}
\begin{aligned}
M_t = \mathcal W(R_{t+1},f_t),\quad
C_{t,\delta}
=\mathcal W\left(R_{t+1},\varphi(f_t)+\delta\right),
\end{aligned}
\tag{2}
\end{equation}
where $\varphi(f_t)=f_t-\operatorname{round}(f_t)$ is the subpixel flow component, $\operatorname{round}(\cdot)$ rounds each component to the nearest integer, and $\delta\in\{(0,0),(r,0),(-r,0),(0,r),(0,-r)\}$ specifies coordinate offsets with integer radius $r$. Thus, $M_t$ samples the next residual along estimated motion, whereas $C_{t,\delta}$ samples near the original image coordinates. Retaining $\varphi(f_t)$ and applying integer offsets gives both references identical bilinear interpolation weights at different locations, avoiding discrepancies due to unmatched fractional sampling.

We compare each reference with the source residual $R_t$ through correlation and distance responses:
\begin{equation}
\begin{aligned}
E_{t,\delta}
&=\operatorname{Avg}_5\left(
\frac{R_t^2+M_t^2+C_{t,\delta}^2}{3}
\right)+\varepsilon,\\
U_{t,\delta}
&=\frac{R_t\odot(M_t-C_{t,\delta})}{E_{t,\delta}},\\
D_{t,\delta}
&=\frac{(R_t-M_t)^2-(R_t-C_{t,\delta})^2}
{E_{t,\delta}},
\end{aligned}
\tag{3}
\end{equation}
where $\odot$ denotes element-wise multiplication, $\varepsilon$ is a small constant for numerical stability, and $\operatorname{Avg}_5$ is a $5\times5$ averaging filter. The shared local energy $E_{t,\delta}$ normalizes the correlation difference $U_{t,\delta}$ and squared-distance difference $D_{t,\delta}$. This normalization can reduce sensitivity to residual magnitude, allowing comparisons across regions with different texture intensities. Forward--backward flow consistency suppresses unreliable correspondences:
\begin{equation}
\begin{aligned}
\rho_t(\xi)
&=\exp\left(
-\frac{
\|f_t(\xi)+\mathcal W(b_t,f_t)(\xi)\|_2^2
}{
1+0.01\|f_t(\xi)\|_2^2
}
\right),\\
\Phi_{t,\delta}
&=3\nu_{t,\delta}\rho_t
\tanh\left(
\frac{[U_{t,\delta};D_{t,\delta}]}{3}
\right),
\end{aligned}
\tag{4}
\end{equation}
where $b_t$ is backward optical flow and $\nu_{t,\delta}$ is a binary mask excluding invalid sampling locations and a three-pixel boundary. The weight $\rho_t$ attenuates responses where the flow estimates disagree, yielding the residual response field $\Phi_{t,\delta}$. We concatenate responses across five offsets and both temporal directions, then apply spatial pooling and a lightweight 3D convolutional encoder to obtain $Z_{t,j}\in\mathbb R^{128}$ for frame pair $t$ and region $j$.

To account for local content, a frozen DINOv2 encoder $F_{teacher}$ provides $P=16$ regional descriptors of dimension 384 per frame. We average adjacent-frame descriptors and use them to modulate the encoded residual features:
\begin{equation}
\begin{aligned}
K_{t,j}
&=\frac12\left(
F_{teacher}(I_t)_j+
F_{teacher}(I_{t+1})_j
\right),\\
e_{t,j}
&=Z_{t,j}\odot
\left[1+\frac12\tanh(W_cK_{t,j}+b_c)\right],
\end{aligned}
\tag{5}
\end{equation}
where $F_{teacher}(I_t)_j$ is the feature of region $j$, $K_{t,j}$ is its adjacent-frame content descriptor, and $W_c\in\mathbb R^{128\times384}$ and $b_c\in\mathbb R^{128}$ are learnable. The bounded modulation produces MCR evidence $e_{t,j}$ that accounts for local content while retaining its dependence on correspondence measurements.

\subsection{Real-Conditioned Score Adaptation}
\label{sec:rcsa}
Complementing MCR's spatial correspondence, RCSA models contextual consistency through a conditional denoising reference learned from authentic videos (Figure.~\ref{fig:intro}(b)). From global frame views $G=\{G_t\}_{t=1}^{T}$, we construct four quality views $\{\mathcal Q_v(G)\}_{v=0}^{3}$: original, blurred, resized, and quantized. A frozen DINOv2 encoder $F_{teacher}$ and its adaptable counterpart $F_{student}$ produce $H^v=F_{teacher}(\mathcal Q_v(G))$ and $S^v=F_{student}(\mathcal Q_v(G))$, both in $\mathbb R^{T\times P\times d}$. They share the first ten frozen Transformer blocks, while the student's final two blocks and output normalization are trainable.

We build the conditional reference in the frozen feature space by projecting regional features into a compact space and pooling their surrounding context:
\begin{equation}
\begin{aligned}
z^v_{t,j}
&=B^\top\operatorname{LN}(H^v_{t,j}),\\
c_{t,j}
&=\left[
\frac{1}{P-1}\sum_{k\ne j}z^0_{t,k};\
\frac{1}{TP-1}\sum_{(u,k)\ne(t,j)}z^0_{u,k}
\right],
\end{aligned}
\tag{6}
\end{equation}
where $\operatorname{LN}$ is layer normalization without affine parameters and $B\in\mathbb R^{d\times d_z}$ is a fixed orthogonal projection with $d_z=64$. Each of the $T$ sampled frames contains $P=16$ regions. The projected feature is $z^v_{t,j}$, and $c_{t,j}\in\mathbb R^{2d_z}$ pools original-view features over other regions in the current frame and all other $(u,k)$ pairs in the video, excluding the target $(t,j)$.

A conditional noise predictor $\epsilon_\theta(\widetilde z,c,\sigma)$ learns to estimate the Gaussian noise perturbing a feature at scale $\sigma$. It is trained exclusively on authentic videos:
\begin{equation}
\mathcal L_{\mathrm{den}}
=\mathbb E_{\substack{X\sim p_{\mathrm{real}}\\v,t,j,\sigma,\eta}}
\left[
\frac{1}{d_z}
\left\|
\epsilon_\theta(z^v_{t,j}+\sigma\eta,c_{t,j},\sigma)
-\eta
\right\|_2^2
\right],
\tag{7} 
\end{equation}
where $p_{\mathrm{real}}$ is the authentic training distribution, $v$ is sampled uniformly over the four quality views, $(t,j)$ indexes the sampled region, and $\eta\sim\mathcal N(0,I_{d_z})$, with $I_{d_z}$ the identity matrix. The noise scale is sampled from $\Sigma=\{0.05s,0.15s,0.4s\}$, with $s=2$ by default. Under optimal noise prediction, $-\epsilon_\theta(\widetilde z,c,\sigma)/\sigma$ estimates the score of the Gaussian-smoothed conditional feature distribution, providing a reference for evaluating variations relative to authentic content.

We probe this reference at each noise scale using deterministic symmetric perturbations along a fixed alternating-sign direction $u\in\{-1,1\}^{d_z}$. Omitting region indices:
\begin{equation}
\begin{aligned}
\epsilon^\pm_{v,\sigma}
&=\epsilon_\theta(z^v\pm\sigma u,c,\sigma),\\
a_{v,\sigma}
&=-\frac{\sigma}{2}(\epsilon^+_{v,\sigma}+\epsilon^-_{v,\sigma}),\\
b_{v,\sigma}
&=\frac12(\epsilon^+_{v,\sigma}-\epsilon^-_{v,\sigma})-u.
\end{aligned}
\tag{8}
\end{equation}
Here, $\epsilon^\pm_{v,\sigma}$ are the predictions for positive and negative perturbations. The response $a_{v,\sigma}$ measures the average denoising displacement of the two perturbed features, while $b_{v,\sigma}$ captures the negative directional finite difference of the denoising map $z-\sigma\epsilon_\theta(z,c,\sigma)$. Together, they describe the common denoising correction and its variation along the probing direction.

We concatenate responses over noise scales as $\psi_v=[a_{v,\sigma};b_{v,\sigma}]_{\sigma\in\Sigma}\in\mathbb R^{384}$. Their mean $\mu=\frac14\sum_{v=0}^{3}\psi_v$ and channelwise standard deviation $\omega=\sqrt{\frac14\sum_{v=0}^{3}(\psi_v-\mu)^2+10^{-6}}$ capture common responses and variations across quality views. A decoder uses the aggregate $\mathbf r=[\mu;\omega]\in\mathbb R^{768}$ to adapt global features:
\begin{equation}
\begin{aligned}
q^{(0)}_{t,j}
=&\ H^0_{t,j}
+\frac14\sum_{v=0}^{3}(S^v_{t,j}-H^v_{t,j}) \\
&+\frac12\tanh\left(F_\phi(\operatorname{sg}(\mathbf r_{t,j}))\right),
\end{aligned}
\tag{9}\label{eq:9}
\end{equation}
where $F_\phi:\mathbb R^{768}\rightarrow\mathbb R^{384}$ is learnable and $\operatorname{sg}$ stops gradients. Thus, $q^{(0)}_{t,j}$ combines the original frozen feature, adaptation across quality views, and a correction from the real-conditioned reference. Classification updates $F_{student}$ and $F_\phi$, while $\epsilon_\theta$ receives only $\mathcal L_{\mathrm{den}}$, keeping the reference independent of discriminative supervision.

\subsection{Reversible Evidence Exchange}
REE integrates MCR's local residual evidence with RCSA's global features. To reconcile their different spatial views, we average local evidence as $\bar e_t=\frac1P\sum_{j=1}^{P}e_{t,j}$ and combine adjacent-pair responses for interior frames as $\widetilde e_t=\frac12(\bar e_{t-1}+\bar e_t)$. At the boundaries, $\widetilde e_1=\bar e_1$ and $\widetilde e_T=\bar e_{T-1}$. A linear projection $W_m\in\mathbb R^{d\times128}$ yields $p^{(0)}_{t,j}=W_m\widetilde e_t$, shared across the $P$ global regions of frame $t$.

Starting from $(q^{(0)}_{t,j},p^{(0)}_{t,j})$, REE exchanges evidence through $L$ reversible layers. Each layer $\ell$ uses independently parameterized scalar potentials $V_\ell$ and $K_\ell$ to determine update directions, both of the form
\begin{equation}
\mathcal P_\ell(x)=0.05\|x\|_2^2+
\sum_{k=1}^{m}a_{\ell k}
\log\cosh(\bar w_{\ell k}^{\top}x+b_{\ell k}),
\tag{10}  
\end{equation}
where $\mathcal P_\ell$ denotes either potential, $x\in\mathbb R^d$, and $m=96$ is the number of nonlinear terms. We set $\bar w_{\ell k}=w_{\ell k}/\|w_{\ell k}\|_2$ and $a_{\ell k}=\frac12\tanh(\alpha_{\ell k})$, with learnable $w_{\ell k}$, $b_{\ell k}$, and $\alpha_{\ell k}$. Omitting temporal and regional indices, each layer performs three alternating updates:
\begin{equation}
\begin{aligned}
p^{(\ell+\frac12)}
&=p^{(\ell)}-\frac{h_\ell}{2}\nabla V_\ell(q^{(\ell)}),\\
q^{(\ell+1)}
&=q^{(\ell)}+h_\ell\nabla K_\ell(p^{(\ell+\frac12)}),\\
p^{(\ell+1)}
&=p^{(\ell+\frac12)}-\frac{h_\ell}{2}\nabla V_\ell(q^{(\ell+1)}),
\end{aligned}
\tag{11}\label{eq:11}
\end{equation}
where $\nabla$ is the gradient with respect to the input state and $h_\ell=\frac12\operatorname{sigmoid}(\tau_\ell)$ is a learnable step size. These updates couple the evidence streams through invertible gradient shears. The layer Jacobian $J_\ell$ satisfies $J_\ell^\top\Omega J_\ell=\Omega$, where $\Omega=\left[\begin{smallmatrix}0&I_d\\-I_d&0\end{smallmatrix}\right]$ is the standard symplectic matrix and $I_d$ is the identity matrix. The exchange therefore preserves the joint representation's volume and recoverability.

After $L=3$ layers, we concatenate $[q^{(L)}_{t,j};p^{(L)}_{t,j}]$, project to 128 dimensions, and add learnable spatiotemporal positional embeddings. A two-layer Transformer encoder produces $N=TP$ tokens $\{z_n^{\mathrm{out}}\}_{n=1}^{N}$. We compute attention weights $\beta_n=\operatorname{softmax}_n(w_a^\top z_n^{\mathrm{out}}+b_a)$, with learnable $w_a$ and $b_a$, and the token mean $\bar z=\frac1N\sum_n z_n^{\mathrm{out}}$. The prediction combines attention pooling with channelwise variation:
\begin{equation}
\begin{aligned}
v_{\mathrm{out}}
&=\sqrt{\frac1N\sum_{n=1}^{N}(z_n^{\mathrm{out}}-\bar z)^2},\\
\widehat y
&=\operatorname{sigmoid}\left(
g_\gamma\left(
\left[\sum_{n=1}^{N}\beta_n z_n^{\mathrm{out}};v_{\mathrm{out}}\right]
\right)\right),
\end{aligned}
\tag{12}
\end{equation}
where $v_{\mathrm{out}}$ is the channelwise standard deviation across tokens, $g_\gamma$ is the classification head, and $\widehat y$ is the probability of an AI-generated video. Attention pooling emphasizes informative features, while the standard deviation retains spatiotemporal variation.

\begin{table*}[t]
\centering
\caption{Comparison with existing AI-generated video detectors across six benchmarks. The best and second-best results are highlighted in \textbf{bold} and \underline{underlined}, respectively. Average results are reported as mean $\pm$ standard deviation. $\diamond$ indicates training-free methods. $\dagger$ denotes a significant improvement over the strongest competing method for the corresponding average metric (two-sided paired $t$-test, $p<0.05$).}
\label{tab:main_results}
\begingroup
\footnotesize
\setlength{\tabcolsep}{2.5pt}
\renewcommand{\arraystretch}{1.10}
\begin{tabular}{@{}l*{14}{c}@{}}
\toprule
\multirow{2}{*}{Method}
& \multicolumn{2}{c}{Pair2}
& \multicolumn{2}{c}{AIGVDBench}
& \multicolumn{2}{c}{ComGenVid}
& \multicolumn{2}{c}{GenBuster}
& \multicolumn{2}{c}{ViF-Bench}
& \multicolumn{2}{c}{RA-Bench}
& \multicolumn{2}{c}{Avg.} \\
\cmidrule(lr){2-3}
\cmidrule(lr){4-5}
\cmidrule(lr){6-7}
\cmidrule(lr){8-9}
\cmidrule(lr){10-11}
\cmidrule(lr){12-13}
\cmidrule(l){14-15}
& ACC & AP & ACC & AP & ACC & AP
& ACC & AP & ACC & AP & ACC & AP
& ACC & AP \\
\midrule
ReStraV~\cite{interno2026ai} (NeurIPS'25)
& 72.03 & 73.90 & 63.74 & 58.94
& 52.82 & 68.93 & 87.21 & 18.55
& 50.05 & 48.95 & 50.53 & 51.61
& 62.73$_{\pm 0.72}$ & 53.48$_{\pm 0.89}$ \\

CoCoDetect~\cite{feng2026cocovideo} (CVPR'26)
& 66.65 & 85.12 & 70.08 & 78.90
& 47.49 & 47.64 & 86.39 & 18.71
& 56.14 & 54.44 & 51.50 & 58.63
& 63.04$_{\pm 0.86}$ & 57.24$_{\pm 1.10}$ \\

D3~\cite{zheng2025d3} (ICCV'25)
& 47.79 & 54.91 & 56.49 & 58.86
& 50.84 & 73.59 & \underline{92.21} & 16.62
& 52.18 & 58.25 & 50.02 & 47.00
& 58.25$^{\diamond}$ & 51.54$^{\diamond}$ \\

SPLIT~\cite{hyun2026split} (ECCV'26)
& 64.54 & 63.80 & 45.70 & 47.92
& 44.90 & 39.74 & 9.85 & 8.22
& 45.54 & 43.93 & 49.46 & 48.38
& 43.33$^{\diamond}$ & 42.00$^{\diamond}$ \\

STALL~\cite{ben2026training} (CVPR'26)
& 51.24 & 57.18 & 47.90 & 60.38
& 58.40 & \underline{83.58} & 16.18 & 23.93
& 54.61 & 59.73 & 50.24 & 51.09
& 46.43$^{\diamond}$ & 55.98$^{\diamond}$ \\

STREAM~\cite{zhubeyond} (ICML'26)
& 69.24 & \textbf{92.58} & 67.78 & 82.65
& 62.68 & 82.69 & \textbf{93.65} & 44.64
& 53.04 & 71.92 & 50.64 & 56.41
& 66.17$_{\pm 0.76}$ & 71.81$_{\pm 0.60}$ \\

EFFORT~\cite{yan2024effort} (ICML'25)
& 72.84 & 78.91 & 70.76 & 87.19
& \underline{68.94} & 81.90 & 62.28 & \textbf{53.95}
& 55.28 & \textbf{79.02} & 49.66 & 59.56
& 63.29$_{\pm 0.97}$ & \underline{73.42}$_{\pm 0.47}$ \\

Native Scale~\cite{li2026preserving} (ICLR'26)
& 50.42 & 62.10 & 45.98 & 50.37
& 49.57 & 48.81 & 8.18 & 13.24
& 47.88 & 48.70 & 50.16 & 49.65
& 42.03$_{\pm 1.08}$ & 45.48$_{\pm 1.35}$ \\

NSG-VD~\cite{zhang2026physics} (NeurIPS'25)
& 62.60 & 53.38 & 67.49 & 58.65
& 33.66 & 37.02 & 83.28 & 12.69
& 44.90 & 46.61 & 32.60 & 37.26
& 54.09$_{\pm 1.32}$ & 40.93$_{\pm 1.87}$ \\

DeMamba~\cite{chen2026demamba} (SCIS'26)
& \textbf{75.33} & 87.05
& \textbf{85.78} & \underline{90.55}
& 63.41 & 75.84
& 73.42 & 36.51
& \underline{64.97} & \underline{75.40}
& \underline{53.98} & \underline{62.12}
& \underline{69.48}$_{\pm 0.39}$ & 71.25$_{\pm 0.71}$ \\

\midrule
\textbf{SCRiD (Ours)}
& \underline{73.11} & \underline{88.23}
& \underline{82.38} & \textbf{93.62}
& \textbf{72.94} & \textbf{84.20}
& 85.42 & \underline{46.30}
& \textbf{66.63} & 68.23
& \textbf{55.84} & \textbf{64.81}
& \textbf{72.72}$_{\pm 0.86}^{\dagger}$
& \textbf{74.23}$_{\pm 0.53}^{\dagger}$ \\
\bottomrule
\end{tabular}
\endgroup
\end{table*}

\subsection{Training Objective and Inference}
SCRiD jointly optimizes video classification and authentic-video denoising (Sec.~\ref{sec:rcsa}). For a mini-batch of $N_{\mathrm b}$ videos, $\mathcal L_{\mathrm{cls}}=\frac{1}{N_{\mathrm b}}\sum_{i=1}^{N_{\mathrm b}}\operatorname{BCE}(\widehat y_i,y_i)$, where $\operatorname{BCE}$ is binary cross entropy, $y_i\in\{0,1\}$ is the ground-truth label, and $\widehat y_i$ is the predicted probability of an AI-generated video. The total objective with $\lambda=0.1$ is as follows:
\begin{equation}
\mathcal L=\mathcal L_{\mathrm{cls}}+\lambda\mathcal L_{\mathrm{den}},
\tag{13}
\end{equation}

During training, $F_{teacher}$, the optical flow estimator, and $B$ remain frozen. The classification loss updates $F_{student}$, the MCR encoder and content modulation, the RCSA response decoder, REE, and the classification head. The denoising loss updates only $\epsilon_\theta$. Classification gradients stop at $\mathbf r$. Both objectives are optimized jointly.

At inference, all parameters are fixed. RCSA uses the same four quality views and three noise scales as in training, probing with deterministic symmetric perturbations. REE processes the resulting representations to predict $\widehat y$, with $\widehat y>0.5$ indicating an AI-generated video.

\section{Experiments}
\label{sec:exper}

\subsection{Experimental Setup}
\textbf{Datasets and Evaluation Protocols.} We evaluate SCRiD on 6 benchmarks covering 88 generator or task subsets: GenVidBench~\cite{ni2026genvidbench}, AIGVDBench~\cite{ma2026your}, ComGenVid~\cite{ben2026training}, GenBuster~\cite{wen2025busterx}, ViF-Bench~\cite{li2026skyra}, and RA-Bench~\cite{liang2026can}. Following the GenVidBench-143k protocol, we train exclusively on Pair1 containing authentic videos and synthetic videos from 4 generators, and evaluate on Pair2 with 4 unseen generators. We further assess cross-dataset generalization on AIGVDBench, ComGenVid, GenBuster, and ViF-Bench, and examine detection performance in realistic crisis scenarios on RA-Bench. We report video-level accuracy (ACC) and average precision (AP).



\noindent\textbf{Implementation Details.} We use DINOv2 ViT-S/14~\cite{oquab2023dinov2} as the backbone and train the student's last 2 Transformer blocks while freezing the teacher and shared backbone. We uniformly sample 8 frames per video from an approximately 2-second central segment and construct $224\times224$ global and local views. DIS \cite{kroeger2016fast} is used to estimate Optical flow, and MCR uses a coordinate offset radius of $r=1$ pixel. We train SCRiD with consistent horizontal flipping and mild brightness augmentation for 10 epochs on an NVIDIA L20 GPU using AdamW with a batch size of 16, weight decay of 0.01, and cosine learning rate decay. Learning rates are $3\times10^{-5}$ for the trainable student blocks and $3\times10^{-4}$ for the remaining parameters.

\noindent\textbf{Baselines.} We compare SCRiD with 10 representative methods: DeMamba~\cite{chen2026demamba}, D3~\cite{zheng2025d3}, STALL~\cite{ben2026training}, ReStraV~\cite{interno2026ai}, CoCoDetect~\cite{feng2026cocovideo}, Native Scale~\cite{li2026preserving}, NSG-VD~\cite{zhang2026physics}, STREAM~\cite{zhubeyond}, SPLIT~\cite{hyun2026split}, and EFFORT~\cite{yan2024effort}. These methods span diverse detection strategies, including temporal statistics, spatial and temporal likelihoods, pretrained visual representations, and learning approaches based on different forensic cues.

\subsection{Main Results}
Table~\ref{tab:main_results} presents the comparison of SCRiD with ten representative detectors across six benchmarks. SCRiD achieves the highest average ACC of 72.72\% and AP of 74.23\%, outperforming the strongest baselines by 3.24 and 0.81 percentage points (pp), respectively. These improvements are statistically significant across five runs according to a two-sided paired $t$-test ( $p=0.00038$ for ACC and $p=0.0319$
for AP). The corresponding 95\% confidence intervals for the performance gains are [2.42, 4.06] and [0.11, 1.51] pp, respectively. Its advantage is also reflected in the individual benchmarks, where SCRiD ranks first or second in ten of the twelve metrics. Specifically, SCRiD achieves the best results in both metrics on ComGenVid and RA-Bench by improving ACC by 4.00 and 1.86 pp over the strongest competing methods, respectively. It also obtains the highest AP of 93.62\% on AIGVDBench and the highest ACC of 66.63\% on ViF-Bench. Notably, the improvements extend across benchmarks with different generation sources and video distributions, demonstrating the strong generalization capability of SCRiD. Furthermore, SCRiD maintains competitive performance even when ACC and AP exhibit different trends. For instance, although DeMamba achieves higher ACC on AIGVDBench, SCRiD surpasses it by 3.07 pp in AP. Overall, these results demonstrate that incorporating spatial correspondence and contextual consistency enables SCRiD to capture transferable detection evidence across diverse unseen generators and video content.

\subsection{Ablation Study}

\begin{table}[t]
\centering
\caption{Component ablation of SCRiD. Average ACC and AP (\%) across 6 benchmarks.}
\label{tab:component_ablation}
\footnotesize
\begin{tabular}{ccccc}
\toprule
MCR & RCSA & REE & ACC & AP \\
\midrule
-- & -- & --
& 61.22\loss{11.50} & 63.62\loss{10.61} \\

$\checkmark$ & -- & --
& 68.58\loss{4.14} & 70.09\loss{4.14} \\

-- & $\checkmark$ & --
& 64.14\loss{8.58} & 66.06\loss{8.17} \\

$\checkmark$ & $\checkmark$ & --
& 69.91\loss{2.81} & 71.22\loss{3.01} \\

\midrule
$\checkmark$ & $\checkmark$ & $\checkmark$
& \textbf{72.72} & \textbf{74.23} \\
\bottomrule
\end{tabular}
\end{table}

\begin{table}[t]
\centering
\caption{Ablation of key design choices in SCRiD. Average ACC and AP (\%) across six benchmarks.}
\label{tab:design_ablation}
\footnotesize

\begin{tabularx}{\columnwidth}{@{}>{\raggedright\arraybackslash}Xcc@{}}
\toprule
Variant & ACC & AP \\
\midrule

\multicolumn{3}{@{}l}{\textit{MCR}} \\
Flow aligned residuals replacing $\Phi_{t,\delta}$
& 69.64\loss{3.08} & 72.36\loss{1.87} \\

Remove content modulation ($e_{t,j}=Z_{t,j}$)
& 65.81\loss{6.91} & 70.11\loss{4.12} \\

\addlinespace
\multicolumn{3}{@{}l}{\textit{RCSA}} \\
Freeze randomly initialized $\epsilon_\theta$
& 68.66\loss{4.06} & 72.61\loss{1.62} \\

Allow $\mathcal{L}_{\mathrm{cls}}$ to update $\epsilon_\theta$
& 72.14\loss{0.58} & 69.86\loss{4.37} \\

\addlinespace
\multicolumn{3}{@{}l}{\textit{REE}} \\
Parameter-matched residual MLP exchange
& 71.00\loss{1.72} & 71.91\loss{2.32} \\

Reversible coupling w/o symplecticity
& 68.17\loss{4.55} & 69.42\loss{4.81} \\

\midrule
SCRiD (Ours)
& \textbf{72.72} & \textbf{74.23} \\
\bottomrule
\end{tabularx}
\end{table}

To investigate the contributions of SCRiD, we conduct two groups of ablation studies. The first evaluates the individual and combined effects of MCR, RCSA, and REE, while the second examines six variants that modify one key design at a time. For MCR, one variant replaces the matched coordinate response $\Phi_{t,\delta}$ with conventional flow-aligned residuals while retaining the subsequent encoder. The other removes content modulation by setting $e_{t,j}=Z_{t,j}$. For RCSA, one variant keeps the noise predictor $\epsilon_\theta$ randomly initialized without denoising training. The other removes the stop gradient operation in Eq.~\eqref{eq:9}, allowing $\mathcal{L}_{\mathrm{cls}}$ to update $\epsilon_\theta$. For REE, one variant replaces the updates in Eq.~\eqref{eq:11} with a parameter-matched residual MLP. The other adopts reversible coupling without enforcing the symplectic condition $J_\ell^\top\Omega J_\ell=\Omega$.  

\subsection{Robustness Evaluation}
\begin{table*}[t]
\centering
\caption{Robustness evaluation on ViF-Bench, ComGenVid, and GenBuster. ACC (\%) is averaged across the three benchmarks. $\Delta$ denotes the mean accuracy drop relative to clean inputs in percentage points (pp).}
\label{tab:robustness}
\footnotesize
\begin{tabular}{lccccccccccc}
\toprule
& & \multicolumn{8}{c}{Corrupted inputs} & \multicolumn{2}{c}{Summary} \\
\cmidrule(lr){3-10}
\cmidrule(lr){11-12}
Method & Clean & Blur & Resize & Noise & $\gamma=0.8$ & $\gamma=1.2$ & Sat. & Cont. & JPEG & Mean & $\Delta$ \\
\midrule
\textbf{SCRiD (Ours)} & 75.00 & 74.52 & 74.04 & 74.37 & 74.57 & 74.63 & 73.85 & 74.35 & 73.96 & \textbf{74.24} & 0.76 \\
ReStraV & 63.36 & 62.64 & 63.25 & 63.59 & 63.12 & 63.30 & 63.09 & 62.80 & 63.41 & 63.14 & \textbf{0.22} \\
CoCoDetect & 63.34 & 60.14 & 62.44 & 63.70 & 63.70 & 63.41 & 63.05 & 60.55 & 62.01 & 62.21 & 1.13 \\
STREAM & 69.79 & 68.51 & 72.17 & 69.98 & 68.91 & 70.19 & 68.70 & 67.04 & 68.92 & 69.27 & 0.52 \\
EFFORT & 62.17 & 59.56 & 60.60 & 68.52 & 61.08 & 61.71 & 61.40 & 64.19 & 50.92 & 60.94 & 1.23 \\
\bottomrule
\end{tabular}
\end{table*}

\begin{figure*}[t]
\centering
\includegraphics[width=0.95\textwidth]{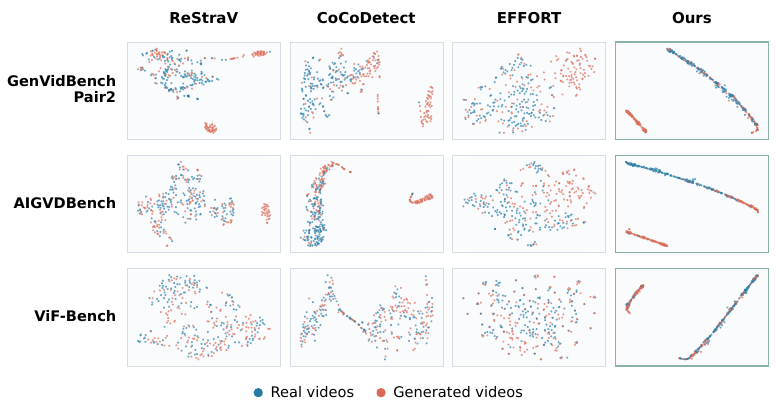}
\caption{t-SNE visualization of video representations extracted by four detection methods on GenVidBench Pair2, AIGVDBench, and ViF-Bench.}
\label{fig:feature_tsne}
\end{figure*}

Table~\ref{tab:component_ablation} evaluates the contributions of the three components. Compared with the baseline, MCR improves the average ACC and AP by 7.36 and 6.47 pp, respectively, while RCSA yields gains of 2.92 and 2.44 pp. Combining both components further increases ACC and AP to 69.91\% and 71.22\%, indicating that spatial correspondence and contextual consistency provide complementary detection evidence. Introducing REE improves the results to 72.72\% ACC and 74.23\% AP, bringing additional gains of 2.81 and 3.01 pp. These improvements demonstrate the benefit of explicitly exchanging information between the two evidence streams.

Table~\ref{tab:design_ablation} further examines the key design choices. For MCR, replacing matched coordinate responses with conventional flow-aligned residuals reduces ACC and AP by 3.08 and 1.87 pp, respectively, while removing content modulation leads to larger drops of 6.91 and 4.12 pp. These results highlight the importance of spatial reference comparisons and content guidance in residual modeling. For RCSA, freezing the randomly initialized noise predictor decreases ACC and AP by 4.06 and 1.62 pp, demonstrating the contribution of the learned conditional denoising reference. Allowing classification gradients to update the noise predictor further reduces AP by 4.37 pp, despite a relatively small ACC drop of 0.58 pp. This suggests that keeping the authentic video reference independent of classification supervision is important for preserving its effectiveness. For REE, replacing the symplectic updates with a parameter-matched residual MLP reduces ACC and AP by 1.72 and 2.32 pp, while adopting reversible coupling without symplecticity causes larger drops of 4.55 and 4.81 pp. These findings support the effectiveness of the proposed structured evidence exchange beyond conventional feature interaction and reversibility alone.

To evaluate the robustness of SCRiD, we apply 7 types of postprocessing perturbations to videos from ViF-Bench, ComGenVid, and GenBuster, including Gaussian blur ($\sigma=1$), resizing ($\times 0.5$), Gaussian noise ($\sigma=2$), gamma correction ($\gamma=0.8$ and $1.2$), saturation and contrast adjustments (both $\times 0.5$), and JPEG compression ($Q=75$). We report the average ACC across the three benchmarks and the mean accuracy drop relative to clean videos, treating the two gamma settings as one perturbation type.
Table~\ref{tab:robustness} shows the robustness evaluation results. SCRiD achieves the highest ACC across all perturbation settings, maintaining an average of 74.24\% compared with 75.00\% on clean videos. Its mean performance drop is only 0.76 pp, with the largest reduction of 1.15 pp occurring under saturation adjustment. Although ReStraV and STREAM exhibit smaller average drops of 0.22 and 0.52 pp, their absolute ACC values remain substantially lower than those of SCRiD. In contrast, CoCoDetect suffers a 3.20 pp drop under Gaussian blur, while EFFORT experiences a pronounced reduction of 11.25 pp under JPEG compression. These results demonstrate that SCRiD preserves strong detection accuracy with limited degradation under diverse post-processing operations.

\subsection{Visualization Analysis}

Figure~\ref{fig:feature_tsne} presents the t-SNE visualization of video representations extracted by four detection methods across 3 benchmarks. Compared with the relatively scattered feature distributions of existing methods, SCRiD produces more coherent and structured representations, with samples forming structured feature distributions across the benchmarks. Notably, on GenVidBench Pair2 and AIGVDBench, generated videos form distinct local clusters separated from the main feature distributions. These observations suggest that incorporating spatial correspondence and contextual consistency helps SCRiD learn more organized video representations that capture differences between authentic and generated content.


Due to space limitations, more experiments and analysis are provided in the Supplementary Material, including \textbf{Computational Efficiency (Sec. B)}, \textbf{Hyperparameter Analysis (Sec. C)}, \textbf{Visualization Analysis on Denoising Reference and Feature Similarity under Blur (Sec. D)}, and \textbf{Qualitative Mechanism Analysis of MCR, RCSA, and REE (Sec. E)}. Please refer to the Supplementary Material for more details.

\section{Conclusion}
\label{sec:conclusion}
In this work, we propose SCRiD, a reference-informed framework for generalizable AI-generated video detection that interprets video discrepancies through spatial correspondence and contextual consistency. Specifically, we introduce MCR to characterize temporal residual associations under motion and coordinate references, and develop RCSA to learn a conditional denoising reference from authentic videos for capturing content-dependent feature variations. We further design REE to integrate these complementary representations through reversible symplectic evidence exchange. Extensive experiments across 6 benchmarks demonstrate the strong generalization capability of SCRiD, achieving average ACC and AP of 72.72\% and 74.23\%, respectively. Ablation studies validate the contributions of the proposed components, while robustness evaluations demonstrate stable detection performance under common post-processing operations. Overall, these results highlight the effectiveness of interpreting video variations through complementary references for detecting videos generated by unseen methods.

\section*{LLM usage statement}
We used OpenAI's ChatGPT to assist with drafting and revising portions of the manuscript, including language polishing and improving clarity. ChatGPT was also used to support experimental iteration, including discussions of experimental configurations and implementation issues. The authors developed the methodology, implemented and conducted the experiments, and independently analyzed the results and formulated the scientific conclusions.

\balance
{\small
\bibliographystyle{ieeenat_fullname}
\bibliography{main}
}

\clearpage
\nobalance
\twocolumn[{
  \begin{center}
  {\Large\bfseries Supplementary Material\par}
  \vspace{8pt}
  \end{center}
}]

\appendix

\section{A Brief Introduction to Baselines}
We compare SCRiD with ten representative methods for AI-generated content detection, including
\textbf{DeMamba}~\cite{chen2026demamba},
\textbf{D3}~\cite{zheng2025d3},
\textbf{STALL}~\cite{ben2026training},
\textbf{ReStraV}~\cite{interno2026ai},
\textbf{CoCoDetect}~\cite{feng2026cocovideo},
\textbf{Native Scale}~\cite{li2026preserving},
\textbf{NSG-VD}~\cite{zhang2026physics},
\textbf{STREAM}~\cite{zhubeyond},
\textbf{SPLIT}~\cite{hyun2026split},
and \textbf{EFFORT}~\cite{yan2024effort}.
These methods cover diverse detection strategies, ranging from temporal inconsistency analysis and forensic artifact modeling to pretrained visual representations and training-free detection.
A brief introduction to each baseline is provided as follows:

\begin{itemize}
    \item \textbf{DeMamba}~\cite{chen2026demamba}: It employs Mamba-based sequence modeling on local visual features to capture spatial and temporal inconsistencies in generated videos, while incorporating global representations for classification.

    \item \textbf{D3}~\cite{zheng2025d3}: This training-free method exploits second-order temporal differences in pretrained visual representations to identify irregular dynamics in AI-generated videos.

    \item \textbf{STALL}~\cite{ben2026training}: It combines spatial and temporal likelihood scores derived from pretrained video features and authentic-video statistics, enabling detection without training on generated samples.

    \item \textbf{ReStraV}~\cite{interno2026ai}: Inspired by perceptual straightening, ReStraV measures temporal curvature and stepwise distances in pretrained feature space to characterize differences between real and generated video trajectories.

    \item \textbf{CoCoDetect}~\cite{feng2026cocovideo}: This framework learns discriminative spatiotemporal representations through contrastive learning on semantically aligned real-fake video pairs. A confidence-gated mechanism further employs multimodal language model reasoning for uncertain predictions.

    \item \textbf{Native Scale}~\cite{li2026preserving}: It processes videos at their native spatial resolutions using a vision-language backbone, preserving high-frequency forgery artifacts and spatiotemporal inconsistencies that may be lost during conventional resizing and cropping.

    \item \textbf{NSG-VD}~\cite{zhang2026physics}: Based on probability flow conservation, it characterizes video dynamics using normalized spatiotemporal gradients and measures their distributional discrepancies from authentic videos.

    \item \textbf{STREAM}~\cite{zhubeyond}: STREAM exploits I-frames, motion vectors, and residual information from compressed video bitstreams to efficiently capture spatial and temporal forgery artifacts.

    \item \textbf{SPLIT}~\cite{hyun2026split}: It provides training-free detection by combining two-step temporal roughness and local spatial motion incoherence extracted from frozen patch-level visual representations.

    \item \textbf{EFFORT}~\cite{yan2024effort}: It decomposes pretrained representations into orthogonal subspaces and selectively adapts complementary components to capture forgery patterns while preserving pretrained knowledge.
\end{itemize}

\section{Computational Efficiency}

\begin{table*}[t]
\centering
\caption{Computational efficiency comparison of SCRiD and existing detectors. GFLOPs are measured per video. Latency and end-to-end (E2E) latency are reported in milliseconds. ``--'' denotes training-free methods.}
\label{tab:efficiency}
\footnotesize

\begin{tabular}{lcccccccc}
\toprule
\multirow{2}{*}{Method}
& \multicolumn{5}{c}{Inference}
& \multicolumn{3}{c}{Training} \\
\cmidrule(lr){2-6}
\cmidrule(l){7-9}

& \shortstack{Params $\downarrow$\\(M)}
& \shortstack{GFLOPs $\downarrow$ \\per video}
& \shortstack{Latency $\downarrow$ \\(ms)}
& \shortstack{E2E latency $\downarrow$ \\(ms)}
& \shortstack{Peak memory $\downarrow$ \\(GiB)}
& \shortstack{Time $\downarrow$ \\(h)}
& \shortstack{Throughput $\uparrow$ \\(videos/s)}
& \shortstack{Peak memory $\downarrow$ \\(GiB)} \\
\midrule

D3
& 121.25 & 511.39 & 25.47 & 207.02 & 0.71
& -- & -- & -- \\

STALL
& 303.13 & 1822.36 & 81.96 & 231.73 & 1.28
& -- & -- & -- \\

ReStraV
& 22.06 & 293.93 & 15.51 & 201.52 & 0.20
& 0.89 & 12.07 & 1.09 \\

CoCoDetect
& 33.50 & 325.57 & 25.32 & 141.44 & 0.39
& 23.43 & 10.03 & 10.56 \\

Native Scale
& 668.69 & 3912.71 & 56.94 & 286.26 & 1.45
& 19.47 & 4.45 & 26.48 \\

NSG-VD
& 553.07 & 13726.54 & 306.16 & 655.80 & 3.91
& 75.75 & 2.43 & 5.79 \\

STREAM
& 73.00 & 214.27 & 13.46 & 353.71 & 0.38
& 9.11 & 4.20 & 25.22 \\

SPLIT
& 121.25 & 511.39 & 25.37 & 229.99 & 0.71
& -- & -- & -- \\

EFFORT
& 504.61 & 5184.97 & 111.83 & 182.84 & 2.37
& 5.10 & 11.20 & 14.37 \\

DeMamba
& 125.71 & 294.66 & 25.28 & 311.54 & 1.41
& 26.83 & 6.28 & 26.23 \\

\midrule
SCRiD (Ours)
& 27.22 & 558.96 & 23.76 & 242.72 & 0.26
& 5.89 & 10.76 & 6.83 \\
\bottomrule

\end{tabular}
\end{table*}

We compare the computational efficiency of SCRiD with existing detectors in terms of both inference and training. Table~\ref{tab:efficiency} reports model size, computational cost, latency, and memory usage during inference, together with training time, throughput, and peak memory for trainable methods.

As shown in Table~\ref{tab:efficiency}, SCRiD achieves a favorable balance between detection performance and computational efficiency. Firstly, SCRiD maintains a compact model size of 27.22M parameters, low peak inference memory of 0.26 GiB, and practical inference latency of 23.76 ms. Compared with recent high-performing baselines such as EFFORT and DeMamba, SCRiD is substantially more efficient in both memory usage and inference latency while delivering stronger detection performance. In particular, SCRiD uses far fewer parameters and less inference memory than EFFORT, and it also reduces both inference and end-to-end latency relative to DeMamba.
SCRiD is also efficient during training. It requires 5.89 hours of training, achieves a throughput of 10.76 videos/s, and uses 6.83 GiB peak training memory. This training cost is markedly lower than that of several competitive baselines, including CoCoDetect, Native Scale, NSG-VD, and DeMamba. Overall, these results show that SCRiD not only brings clear performance gains over prior detectors, but also preserves a highly practical efficiency profile, making it attractive for real-world deployment.

\section{Hyperparameter Analysis}

\begin{figure}[t]
    \centering
    \includegraphics[width=0.9\columnwidth]{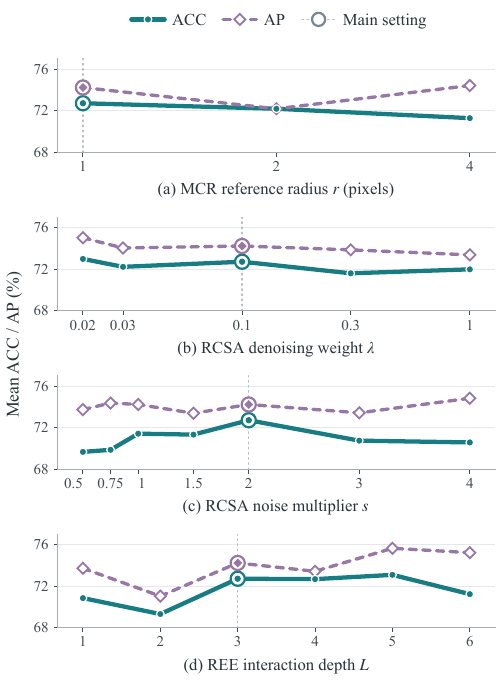}
    \caption{Hyperparameter sensitivity of SCRiD. Results are reported as the mean ACC and AP (\%) over six benchmarks.}
    \label{fig:hyperparameter_analysis}
\end{figure}

\begin{figure}[t]
\centering
\includegraphics[width=\columnwidth]{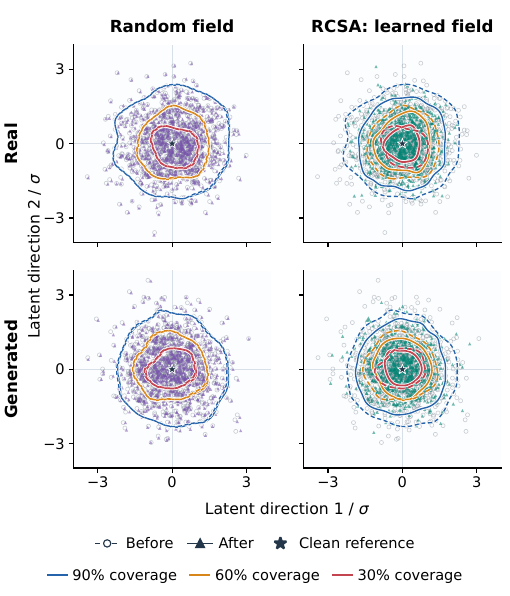}
\caption{Visualization of feature correction before and after RCSA denoising training. The top and bottom rows correspond to authentic and generated videos, respectively. Compared with the randomly initialized predictor (left), the trained predictor (right) moves perturbed features closer to the clean reference.}
\label{fig:rcsa_denoising}
\end{figure}

We analyze the sensitivity of several key hyperparameters in SCRiD. Specifically, we vary one hyperparameter at a time while keeping the others fixed at their default settings. Figure~\ref{fig:hyperparameter_analysis} illustrates the influence of four key hyperparameters. The star markers in Figure~\ref{fig:hyperparameter_analysis} correspond to the adopted default settings in the main experiments. For the coordinate offset radius $r$ in MCR, increasing $r$ from 1 to 4 gradually reduces ACC, while AP exhibits some variation, suggesting that a wider spatial reference does not consistently improve detection. The denoising loss weight $\lambda$ has a relatively limited influence, with both metrics remaining competitive across the tested settings. For the noise scaling factor $s$, the default value $s=2$ achieves the highest ACC of 72.72\%, while both smaller and larger scales generally lead to lower accuracy, indicating the importance of selecting an appropriate noise scale. The number of REE exchange layers $L$ exhibits a nonmonotonic trend. Although $L=5$ achieves the highest ACC and AP of 73.09\% and 75.64\%, respectively, the default setting $L=3$ provides competitive performance with fewer layers. Overall, these results indicate that SCRiD maintains reasonable stability across different hyperparameter settings, with the adopted configuration offering a favorable balance between detection performance and model complexity.

\begin{figure*}[t]
\centering
\includegraphics[width=0.85\textwidth]{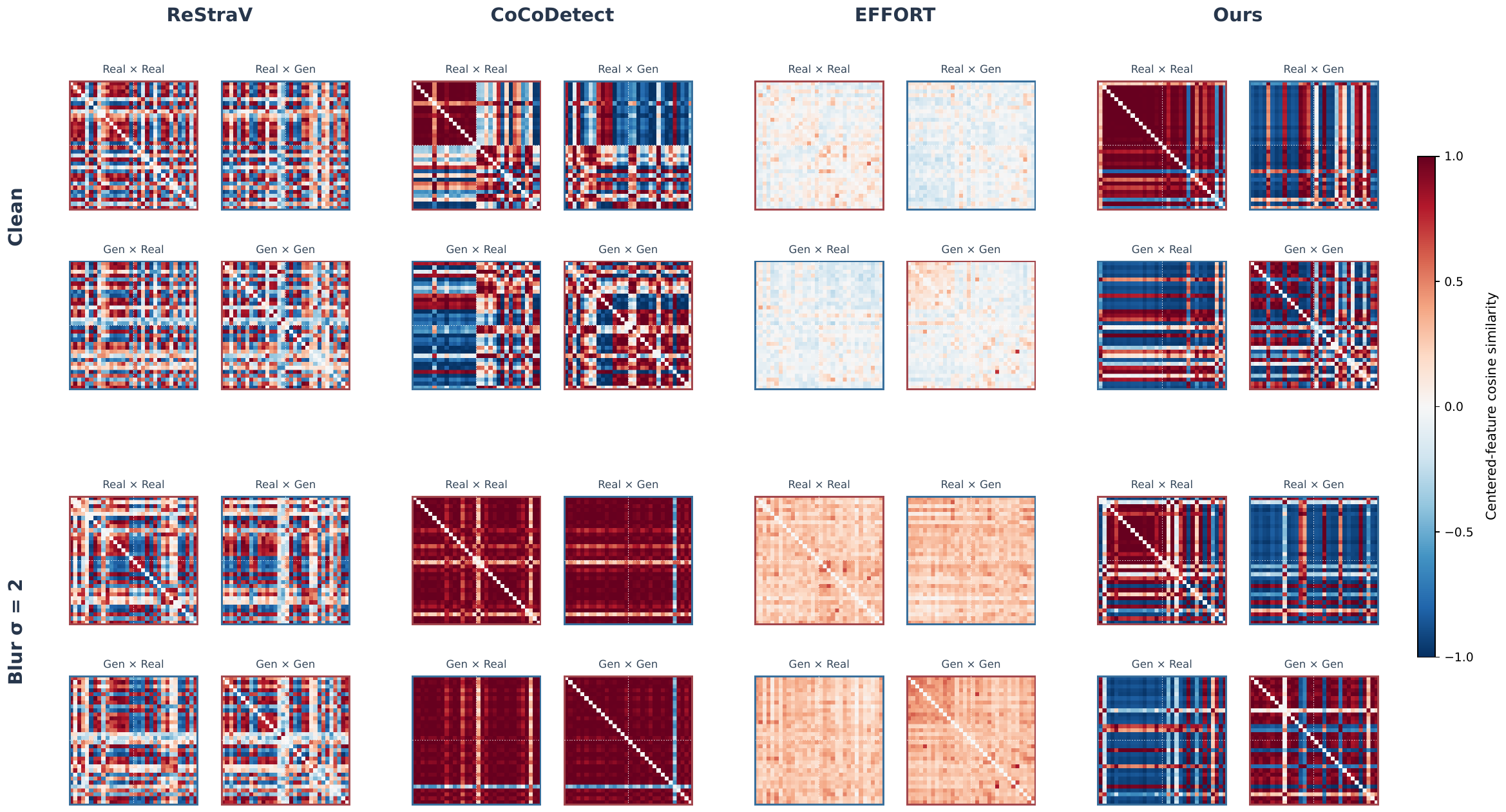}
\caption{Pairwise feature similarity under clean and Gaussian blur ($\sigma=2$) conditions. The top and bottom panels show the similarity matrices before and after blurring, respectively. Each method includes four matrices representing authentic and generated video pairs.}
\label{fig:feature_blur}
\end{figure*}

\section{More Visualization Analysis}


\noindent\textbf{Visualization of Denoising Reference.}
Figure~\ref{fig:rcsa_denoising} visualizes how the conditional noise predictor corrects perturbed features before and after denoising training. The dashed and solid contours represent feature coverage before and after correction, respectively, while the star marks the clean reference. The randomly initialized predictor produces little correction, with the two sets of contours largely overlapping. After training, the corrected features move closer to the clean reference, as reflected by the contraction of the solid contours. This behavior is observed for both authentic and generated videos, indicating that RCSA learns an effective denoising reference that provides a basis for subsequent probing of feature responses.


\noindent\textbf{Feature Similarity under Blur.}
Figure~\ref{fig:feature_blur} compares the pairwise cosine similarities of centered video features before and after Gaussian blurring. Each method is evaluated using four similarity matrices corresponding to authentic and generated video pairs, where red and blue indicate higher and lower similarities, respectively. Under blurring, ReStraV exhibits irregular similarity patterns across both same-class and different-class pairs, making their distinctions less apparent. CoCoDetect and EFFORT show broadly increased similarities across different class combinations, weakening the distinction between authentic and generated features. In comparison, SCRiD maintains relatively high similarities within each class and lower similarities between classes, preserving clearer class-related feature structure under visual degradation.

\section{Qualitative Analysis}

\begin{figure}[t]
\centering
\includegraphics[width=\columnwidth]{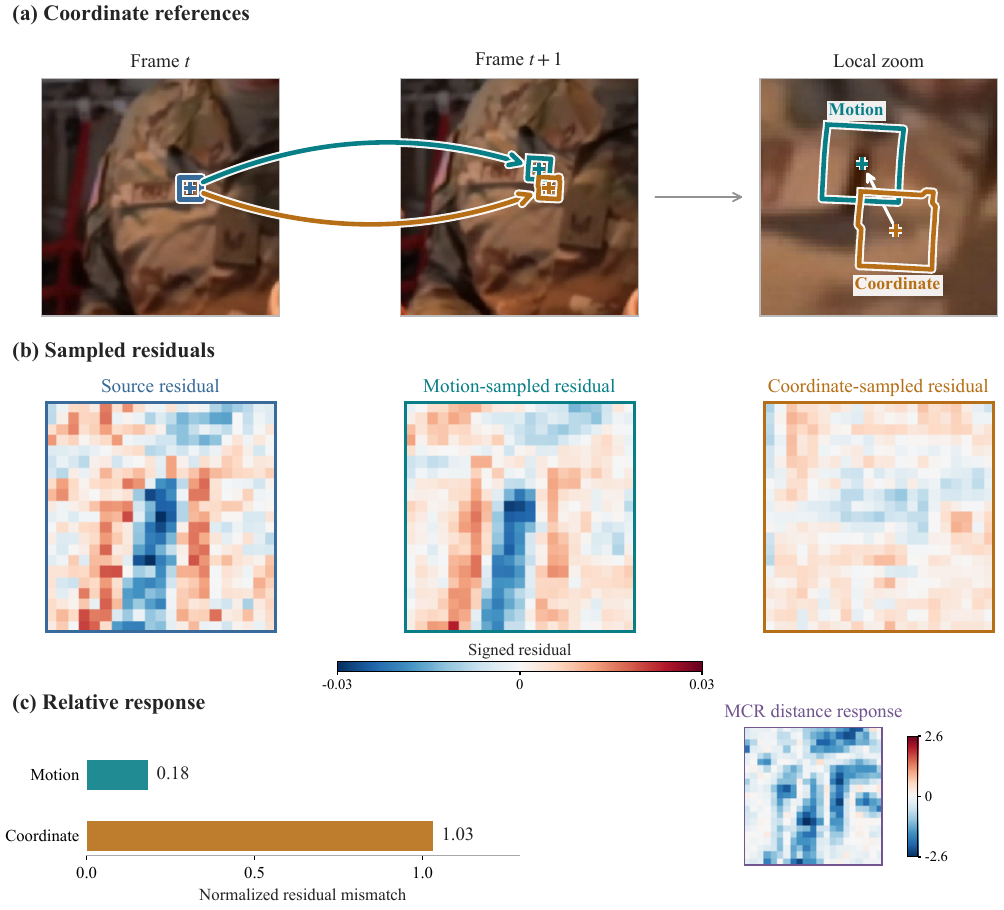}
\caption{Mechanism Analysis of MCR. (a) Motion and coordinate references for an adjacent frame pair. (b) Residual patterns sampled under the two references and the corresponding source residual. (c) Normalized residual mismatches and the resulting MCR distance response.}
\label{fig:mcr_mechanism}
\end{figure}

\begin{figure}[t]
\centering
\includegraphics[width=\columnwidth]{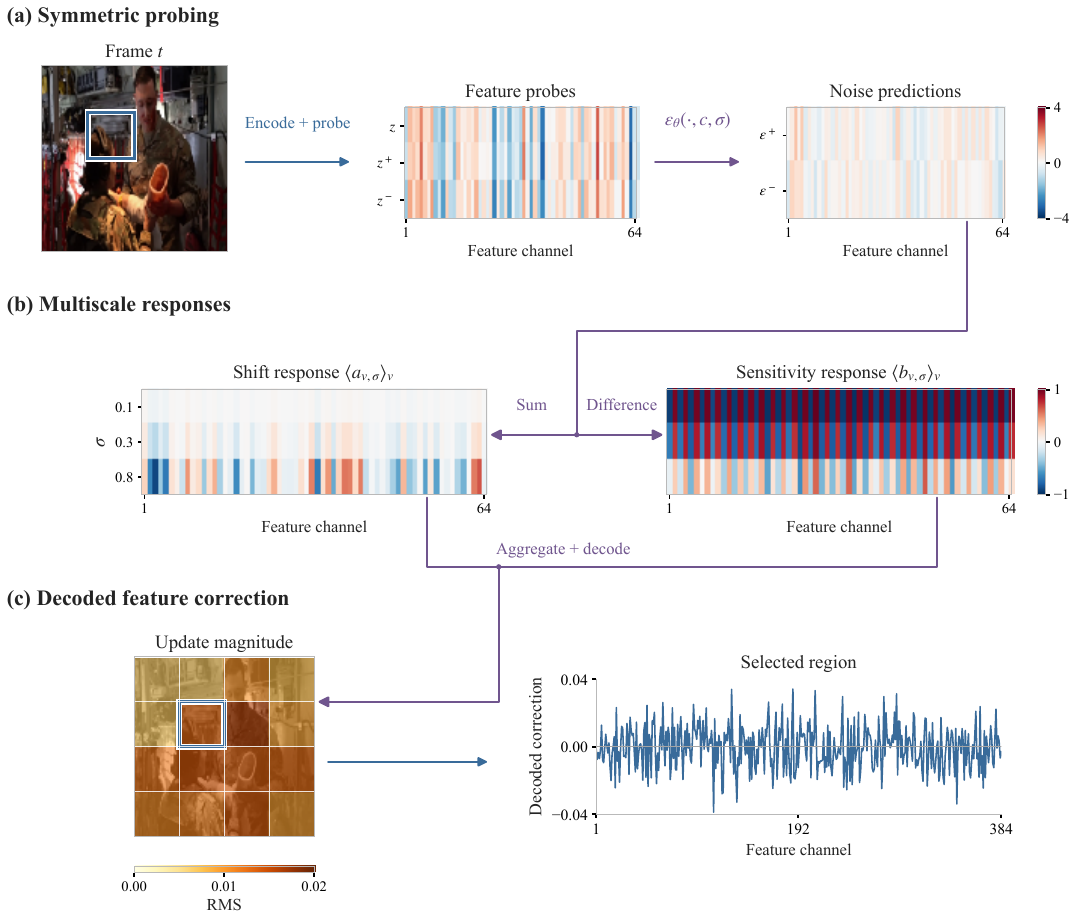}
\caption{Qualitative illustration of feature probing and correction in RCSA. (a) Symmetric perturbations of a selected regional feature and the corresponding conditional noise predictions. (b) Shift and sensitivity responses across three noise scales. (c) Spatial distribution of feature update magnitudes and the channelwise correction for the selected region.}
\label{fig:rcsa_mechanism}
\end{figure}

\begin{figure}[t]
\centering
\includegraphics[width=\columnwidth]{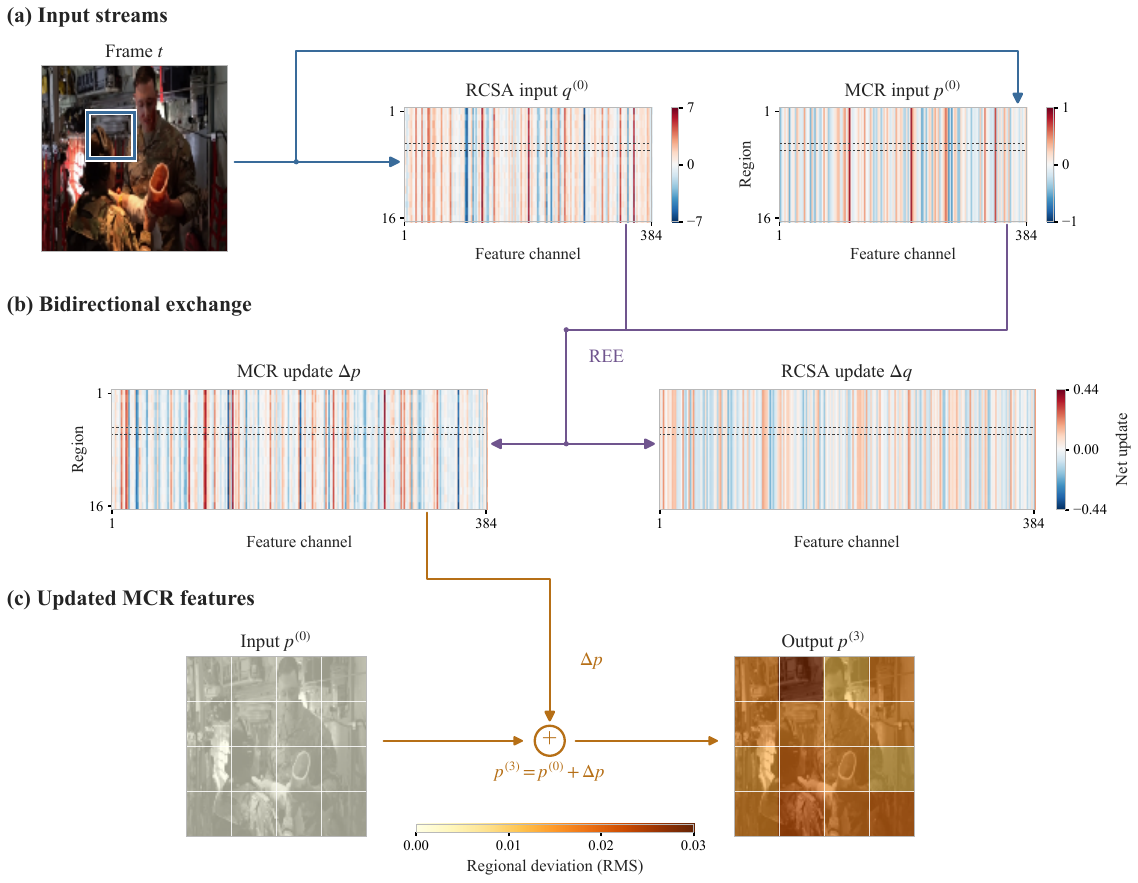}
\caption{Qualitative illustration of evidence exchange in REE. (a) Initial RCSA and MCR representations before interaction. (b) Feature updates in both evidence streams after three reversible exchange layers. (c) Regional variations in the MCR representation before and after evidence exchange.}
\label{fig:ree_mechanism}
\end{figure}

\begin{figure}[t]
\centering
\includegraphics[width=\columnwidth]{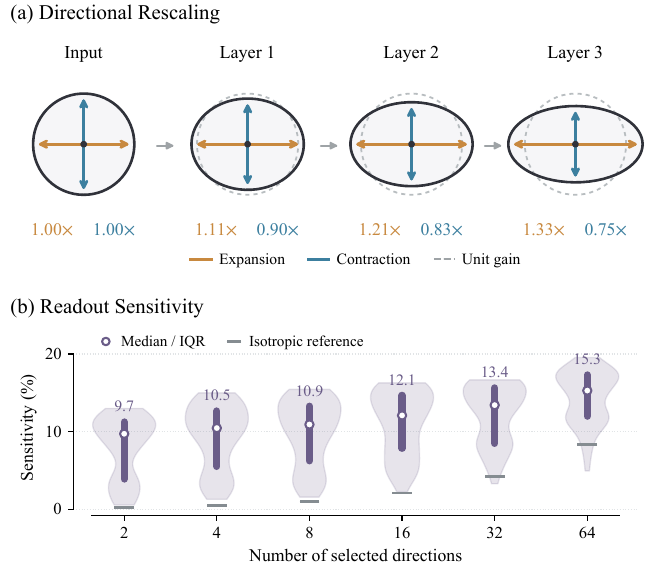}
\caption{Directional effects of Symplectic Updates. (a) Expansion and contraction of feature variations across exchange layers. (b) Classification sensitivity along strongly rescaled directions compared with an isotropic reference.}
\label{fig:symplectic_directional}
\end{figure}

\noindent\textbf{Mechanism Analysis of MCR.}
Figure~\ref{fig:mcr_mechanism} illustrates how MCR compares temporal residual associations under the motion and coordinate references. Starting from the same location in the source frame, the motion reference samples the subsequent residual along the estimated motion, while the coordinate reference samples near the original image coordinates. As shown in the figure, the motion-sampled residual exhibits stronger agreement with the source residual, yielding a normalized mismatch of 0.18 compared with 1.03 under the coordinate reference. This difference is further captured by the MCR distance response, which highlights variations in residual agreement between the two spatial references. These observations demonstrate how MCR converts spatial correspondence information into local temporal evidence.

\noindent\textbf{Mechanism Analysis of RCSA.}
Figure~\ref{fig:rcsa_mechanism} illustrates how RCSA converts conditional denoising responses into feature corrections. For the selected region, symmetric perturbations are applied to its projected feature, and the conditional noise predictor produces corresponding predictions based on the surrounding video context. These predictions are combined to obtain shift and sensitivity responses across three noise scales, capturing the denoising correction and its variation along the probing direction. The aggregated responses are subsequently decoded into feature corrections. As shown in the figure, the correction magnitudes vary across spatial regions, while the selected region exhibits both positive and negative adjustments across feature channels. These observations illustrate how RCSA uses the learned conditional denoising reference to guide the adaptation of visual representations.

\noindent\textbf{Mechanism Analysis of REE.}
Figure~\ref{fig:ree_mechanism} illustrates how REE enables information exchange between the MCR and RCSA evidence streams. Before interaction, the MCR representation is shared across spatial regions within each frame, while the RCSA representation retains regional variations. Through reversible exchange layers, the two coupled states influence each other, producing updates in both representations. As shown in the figure, the initially identical MCR features acquire distinct regional patterns after three exchange layers, reflecting the incorporation of regional information from RCSA. These observations demonstrate how REE facilitates bidirectional evidence interaction while enriching the representations of the two streams.

\noindent\textbf{Mechanism Analysis of Symplectic Updates.}
Figure~\ref{fig:symplectic_directional} further examines the directional behavior of the symplectic updates in REE. While the symplectic condition preserves the volume and recoverability of the joint representation, it allows different directions to undergo distinct transformations. As shown in Figure~\ref{fig:symplectic_directional} (a), feature variations with the same initial magnitude exhibit progressive expansion or contraction across successive exchange layers. To assess their relevance to the final prediction, we measure readout sensitivity by the percentage change in the classification output under equal-magnitude feature perturbations. An isotropic reference is constructed using perturbations of the same magnitude. Figure~\ref{fig:symplectic_directional} (b) further shows that the directions with pronounced rescaling exhibit higher classification sensitivity than this isotropic reference. These observations illustrate how the structured symplectic updates enable nonuniform directional adjustments during evidence exchange, with the strongly affected directions also contributing to the final prediction.

%
%
%
\end{document}